\documentclass{ieeetj}
\usepackage{cite}
\usepackage{amsmath,amssymb,amsfonts}
\usepackage{algorithmic}
\usepackage{graphicx,color}
\usepackage{textcomp}
\usepackage{xcolor}
\usepackage{hyperref}
\usepackage{multirow}
\hypersetup{hidelinks=true}
\usepackage{algorithm,algorithmic}

\def\BibTeX{{\rm B\kern-.05em{\sc i\kern-.025em b}\kern-.08em
    T\kern-.1667em\lower.7ex\hbox{E}\kern-.125emX}}
\AtBeginDocument{\definecolor{tmlcncolor}{cmyk}{0.93,0.59,0.15,0.02}\definecolor{NavyBlue}{RGB}{0,86,125}}

\def\OJlogo{}
\def\seclogo{}

\begin{document}

\title{Differential Attention Unlocks Complementary EEG–Speech Fusion for Emotion Recognition}

\author{Philip H. Lee\authorrefmark{1},
Shreeram Suresh Chandra\authorrefmark{2},
and John H.L. Hansen\authorrefmark{3}}

\affil{\authorrefmark{1}Independent Researcher, USA}
\affil{\authorrefmark{2}Center for Language and Speech Processing (CLSP), Johns Hopkins University, Baltimore, MD, USA}
\affil{\authorrefmark{3}Center for Robust Speech Systems (CRSS), University of Texas at Dallas, Richardson, TX, USA}

\begin{abstract}
Multimodal emotion recognition (MER) increasingly pairs EEG with speech, treating internal neural signals and external vocal expression as informative views of affect. In practice, naive fusion underperforms the stronger single modality, because EEG artifacts inject noise that corrupts the shared representation. We introduce EmoSpeechBrain, a multimodal framework built on the insight that noise suppression is a precondition for effective fusion. Its EEG encoder uses differential attention, taking the difference between two attention maps to cancel shared noise and isolate discriminative neural activity. An attention-based gating adapter aligns both modalities in a shared space and weights each one's contribution to the prediction. On two datasets - PME4 and EAV, EmoSpeechBrain improves MER accuracy by up to 12.9\% over other state-of-the-art (SOTA) EEG encoders, and surpasses unimodal speech and EEG baselines by up to 13.1\% and 23.1\%. These results show that once EEG noise is suppressed, fusion delivers gains that naive combination cannot.
\end{abstract}

\begin{IEEEkeywords}
Multimodal Emotion Recognition, EEG, Speech
\end{IEEEkeywords}


\maketitle

\section{Introduction}
\label{sec:intro}
Emotions shape how people communicate, make decisions, and interact with the world, making their accurate recognition essential for applications ranging from mental health monitoring to human-computer interaction \cite{kiecolt2002emotions}. This has motivated a growing body of work on emotion recognition in a range of modalities, including facial expressions \cite{mollahosseini2017affectnet}, speech \cite{schuller2018speech}, and more commonly, their combination \cite{poria2017review, chudasama2022m2fnet}.

However, most emotion recognition research relies on expressive modalities alone, such as speech \cite{Trigeorgis2016}, facial expression \cite{Xue2021}, and text \cite{Oberlander2018}. These channels capture the outward expression of emotion, which a speaker can shape, suppress, or exaggerate, so they offer only an indirect view of the underlying affective state. EEG provides a physiological signal that is less subject to deliberate control, giving a complementary internal correlate of emotion. This motivates combining EEG with speech for more robust recognition.

\begin{figure*}[t]
    \centering
    \includegraphics[width=0.85\textwidth]{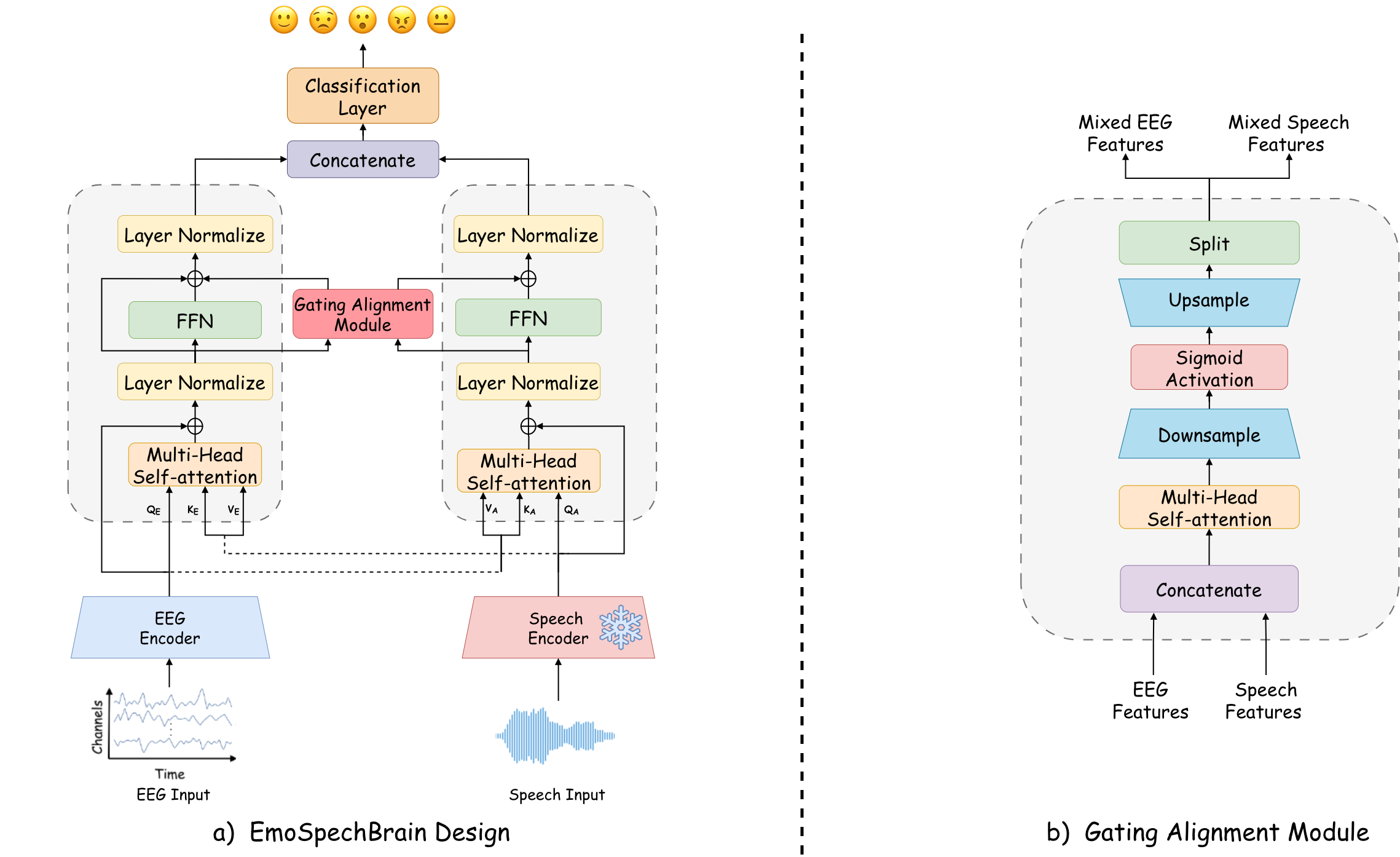}
    \caption{Overall architecture of EmoSpeechBrain. (a) The EEG Encoder
    applies differential attention to isolate discriminative neural activity,
    while a frozen WavLM-Large encoder extracts speech features. The two
    modalities interact through parallel cross-modal attention and are fused by
    the Gating Alignment Module (GAM) before concatenation and classification.
    (b) The GAM concatenates EEG and speech features, refines them with
    self-attention and a sigmoid-gated bottleneck, then splits the result into
    aligned EEG and speech streams, $\hat{Z}_{eeg}$ and $\hat{Z}_{speech}$.}
    \label{fig:framework}
\end{figure*}

Biosignals reveal the physiological side of felt emotion that is not seen in expressive behavior \cite{shu2018review}. Many modalities such as functional magnetic resonance imaging (fMRI) \cite{phan2002functional}, electrodermal activity (EDA) \cite{shu2018review}, and electroencephalograms (EEGs) \cite{alarcao2017emotions, zheng2015investigating} have been explored individually to understand neurophysiolocial factors. Each modality varies considerably in practical utility. For example, fMRI data suffer from hardware costs and strict environmental constraints, while EDA effectively captures sympathetic arousal but cannot reliably distinguish emotional valence \cite{shu2018review}. EEG, on the contrary, offers a noninvasive means of capturing brain activity with semantic correspondence to emotional states.

Despite this, tight integration of EEG and speech for MER remains comparatively underexplored. This is due in part to the challenge of real-world EEG signals, which can carry artifacts such as muscle movement \cite{zhang2021eeg} and eye blinks \cite{sun2020eeg}. Left unaddressed, these artifacts propagate through cross-modal fusion between EEG and speech representations, degrading MER performance by forcing the model to attend to spurious detail rather than emotionally meaningful information. Excluding the EEG modality forfeits non-redundant information, as it captures internal physiological correlates of affect that the speech signal does not convey \cite{Cai2023}. The tradeoff is therefore not whether to use EEG but how: its value comes paired with noise that must be suppressed before fusion can help. We argue that MER with EEG requires addressing both challenges together. The model must first denoise the neural signal by suppressing artifacts from involuntary physiological activity and then fuse the preserved emotional information with speech. %\src{I think one sentence here - about what we refer to as denoising and contextualising it will help}

We address these limitations with EmoSpeechBrain, designed around the following premise: artifact suppression and cross-modal fusion are interdependent, since fusion can only exploit neural features whose discriminative content has been emphasized, and that emphasis is only useful once aligned with speech. Our EEG Encoder integrates differential attention \cite{ICLR2025_00b67df2}, which isolates discriminative neural activity by canceling shared noise across attention maps, with a local feature extraction branch that captures fine-grained temporal structure. The information-rich EEG representations are then fused with speech through a cross-modal Gating Alignment Module (GAM) \cite{qiu2025gated}, whose attention-based gating aligns informative signal rather than the artifacts that would otherwise corrupt the shared representation. The two stages jointly yield a tighter task-relevant EEG-speech coupling and achieve state-of-the-art accuracy on two widely used datasets - EAV \cite{lee2024eav} and PME4 \cite{pme4dataset}.

The key contributions are as follows:
\begin{itemize}
    \item We introduce a noise-aware EEG Encoder whose differential attention subtracts attention patterns common to both maps, reducing the weight placed on artifact-driven activity and isolating discriminative neural activity before fusion.
    %\src{I think especially here - defining what noise means in our context is imp}
    \item We propose the Gating Alignment Module (GAM), a lightweight mechanism that fuses EEG and speech by aligning them in a shared space and gating out emotionally irrelevant information.
    \item We show that the multimodal approach beats unimodal speech and EEG baselines on two well-known datasets, PME4 and EAV, with all gains statistically significant.
    \item We provide a comprehensive ablation study confirming that each component contributes, with differential attention, GAM, and local feature extraction each adding measurable performance.
    \item We perform a frequency ablation showing that differential attention relies less than vanilla attention on the \textit{beta} and \textit{gamma} frequency bands, which are most prone to involuntary movement contamination.
\end{itemize}

\begin{figure*}[t]
    \centering
    \scalebox{0.165}
    {\includegraphics{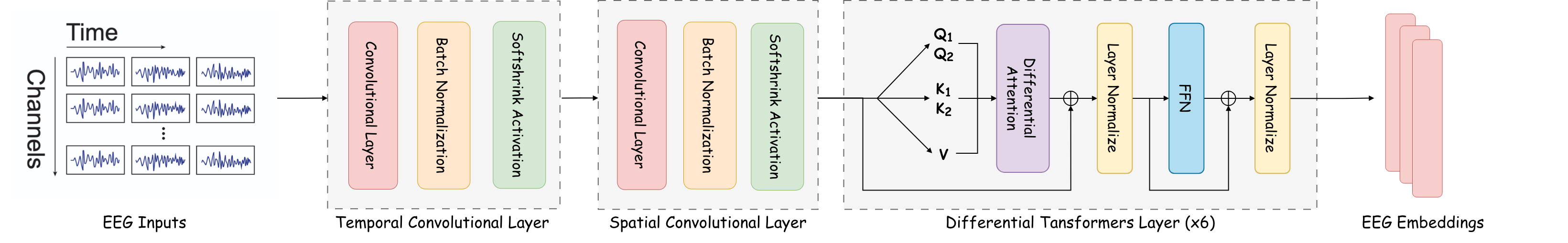}}
    \caption{Proposed EEG encoder architectural design for EmoSpeechBrain.}
    \label{fig:eeg-encoder}
\end{figure*}

\section{Related Works}
\subsection{Common Emotion Recognition Modalities} Early MER systems focused primarily on speech and text, leveraging the complementary strengths of both modalities \cite{zadeh2017tensor}. These approaches demonstrated that the combining modalities consistently outperformed unimodal baselines, with many approaches such as self-supervised learning (SSL) \cite{tsai2019multimodal} achieving SOTA performance. Subsequent work expanded the modality space to include visual expressions, such as facial expressions \cite{mittal2020m3er} alongside speech and text. Video-based MER systems demonstrated further improvements by capturing rich spatial information present in facial expressions alongside acoustic features. However, despite their effectiveness, facial and vocal expressions can be voluntarily suppressed or masked \cite{dsid2026masked}, failing to reflect the true emotional state of a person.

\subsection{Biosignals for Emotion Recognition} To address this concern, researchers have explored brain-related signals as a complementary modality for MER. Among these, fMRI was an early candidate \cite{phan2002functional} due to its detailed spatial resolution of brain activity. However, despite these advantages, fMRI remains impractical for most MER scenarios due to its high cost and cumbersome acquisition setup. More recently, biosignals such as EEG have gained considerable attention as a noninvasive and inexpensive alternative for capturing neural activity that correlates with emotion \cite{alarcao2017emotions, zheng2015investigating}. 

\subsection{Unimodal EEG Emotion Recognition} Unimodal EEG-based emotion recognition has already demonstrated promising results \cite{alarcao2017emotions, zheng2015investigating}, motivating its integration with speech for MER. However, the artifact-ridden nature of EEG signals poses a significant challenge for cross-modal fusion. Our proposed approach addresses this gap through a robust EEG encoder and a lightweight alignment mechanism that enables MER to be more effective.

\section{Main Method}
In this section, we present the main methodology behind EmoSpeechBrain as illustrated in Fig.~\ref{fig:framework}(a). We first discuss our choice for the speech encoder, followed by an in-depth description of layers in the EEG encoder. EmoSpeechBrain also includes an alignment module, GAM to align both modalities. We address the design of GAM, that selects relevant multimodal features and cross-aligns them to EEG and speech.

\subsection{Speech Encoder}
For the speech modality, we select WavLM-Large \cite{chen2022wavlm} as our speech encoder due to its ability to capture long-range dependencies and robustness to noisy conditions \cite{chen2022wavlm}. Given a raw audio input $x \in \mathbb{R}^{L}$, where $L$ is the sequence length, we pass it through a frozen speech encoder $SpeechEncoder(\cdot)$ to extract frame-level embeddings $X_{s} \in \mathbb{R}^{T \times d_{model}}$, where $T$ denotes the number of output frames and $d_{model}$ is the encoder feature dimension. These are then projected through a linear layer to a lower-dimensional representation:
\begin{equation}
\begin{gathered}
    X_{s} = \text{SpeechEncoder}(x) \\[6pt]
    X_{speech} = W_{s} X_{s} + b_{s}.
\end{gathered}
\end{equation}
Here, $W_{s}$ and $b_{s}$ are learnable projection weights and bias, mapping $x_{s}$ to the final speech representation $X_{speech} \in \mathbb{R}^{T \times F}$, with $F$ denoting the final projected feature dimension of the speech modality. Further usage of $X_{speech}$ is explained in Sec.~\ref{sec:bcmal}.

\subsection{EEG Encoder}
The EEG Encoder as shown in Fig.~\ref{fig:eeg-encoder} addresses a fundamental challenge in neural signal processing: raw EEG recordings are contaminated by motion artifacts and background noise. To extract meaningful emotional representations, we adapt the transformer architecture \cite{vaswani2017attention} with targeted modifications to bypass artifact components and isolate the underlying neural signatures of emotion. 

\subsubsection{Local Feature Extraction Layer}
To begin, we perform local feature extraction of the raw EEG input $X \in \mathbb{R}^{L \times C}$, where $L$ is the same sequence length as $X_{speech}$ and $C$ is the number of EEG channels. We pass $X$ through two successive convolutional layers: a \textit{Temporal Convolutional Layer} that captures local patterns across time, followed by a \textit{Spatial Convolutional Layer} that captures dependencies across EEG channels. Each layer is followed by batch normalization and a \textit{softshrink} activation function. This is formally expressed as:\begin{equation}
\begin{gathered}
    X_{T} = \text{Softshrink}(\text{BatchNorm}(\text{TimeConv2D}(X))) \\[6pt]
    X_{ST} = \text{Softshrink}(\text{BatchNorm}(\text{SpaceConv2D}(X_{T}))),
\end{gathered}
\end{equation}

where $X_{T} \in \mathbb{R}^{F \times T \times C}$ denotes the temporally extracted EEG features and $X_{ST} \in \mathbb{R}^{T \times F}$ denotes the final EEG features after spatial filtering. Both $T$ and $F$ match the temporal and feature dimensions of $X_{speech}$. For the convolutional layers, \textit{TimeConv2D(.)} is configured with a kernel size of $(1, k_t)$, where $k_t$ represents the size of the temporal filter, and \textit{SpaceConv2D(.)} is configured with a kernel size of $(k_s, 1)$, where $k_s$ represents the size of the spatial filter. The Softshrink activation function serves as an adaptive noise suppression \cite{feng2025id3rsnet}, mitigating small-magnitude noise temporally and spatially for $X_{T}$ and $X_{ST}$.

\subsubsection{Differential Transformers Layer}
To capture global dependencies, we also apply 6 differential transformer layers, which have been empirically shown to capture more meaningful information from noisy data \cite{ICLR2025_00b67df2}. Following their formulation, the query and key projections of $\hat{X}$ are first split into two halves, $Q_1, Q_2, K_1, K_2 \in \mathbb{R}^{T \times \frac{d}{2}}$, while the value projection $V \in \mathbb{R}^{T \times d}$ remains the same size. This overall projection is expressed as:

\begin{equation}
[Q_1; Q_2] = W_Q\hat{X}, \enspace [K_1; K_2] = W_K\hat{X}, \enspace V = W_V\hat{X},
\end{equation}

where $W_{Q}, W_{K}, W_{V} \in \mathbb{R}^{F \times d}$ are learnable weight matrices. Each query-key pair then computes two separate attention maps, $A_1$ and $A_2$:

\begin{equation}
A_1 = \text{Softmax}\left(\frac{Q_1 K_1^\top}{\sqrt{d/2}}\right), A_2 = \text{Softmax}\left(\frac{Q_2 K_2^\top}{\sqrt{d/2}}\right).
\end{equation} The two maps are then differenced through a learnable scalar $\lambda$, which controls the degree of sparsity in the resulting attention map. This is formulated below:

\begin{equation}
\begin{gathered}
\lambda = \exp(\lambda_{q_1} \cdot \lambda_{k_1}) - \exp(\lambda_{q_2} \cdot \lambda_{k_2}) + \lambda_{\text{init}} \\[6pt]
\mathrm{DiffAtt}(\hat{X}) = (A_1 - \lambda A_2)V,
\end{gathered}
\end{equation}

where $\lambda_{q_1}, \lambda_{k_1}, \lambda_{q_2}, \lambda_{k_2} \in \mathbb{R}^{d}$ are learnable weights and $\lambda_{\text{init}} \in (0,1)$ is an initialization hyperparameter. The difference in $A_1$ and $A_2$ results in an overall sparse attention map that enables EmoSpeechBrain to focus on more relevant concepts. Finally, we extend differential attention to multiple heads as in \cite{vaswani2017attention}, where each head computes differential attention in parallel. This is displayed below: \begin{equation}
\mathrm{MHDA}(\hat{X}) = \left[\mathrm{DiffAtt}_1(\hat{X}), \cdots, \mathrm{DiffAtt}_H(\hat{X})\right]W_O
\end{equation}
where the outputs of each head are concatenated and projected through $W_O$, allowing the model to jointly attend to complementary patterns across time. 

The $MHDA(.)$ output is then passed through layer normalization, a feed-forward network with softshrink activation, and residual connections as formulated below:

\begin{equation}
\begin{gathered}
Z = \mathrm{LayerNorm}(\hat{X} + \mathrm{MHDA}(\hat{X})) \\[6pt]
\mathrm{FFN}(Z) = W_2 \, \mathrm{Softshrink}(W_1 Z + b_1) + b_2 \\[6pt]
X_{eeg} = \mathrm{LayerNorm}(Z + \mathrm{FFN}(Z)).
\end{gathered}
\end{equation} Here, $W_1, W_2 \in \mathbb{R}^{D \times D}$ and $b_1, b_2 \in \mathbb{R}^{D}$ are the learnable weights and biases of the feed-forward network. The residual connections maintain gradient flow while layer normalization stabilizes training, yielding the final EEG embedding $X_{eeg} \in \mathbb{R}^{T \times F}$, which has the exact same dimensional size as $X_{speech}$.

\subsection{Bidirectional Cross-Modal Attention Layer} 
\label{sec:bcmal}
At this stage, we have the output embeddings from both the EEG and speech encoders, $X_{eeg}$ and $X_{speech}$. Both embeddings are passed through a parallel multi-head self-attention sublayer, where the EEG queries attend to speech keys and values and vice versa:
\begin{equation}
\begin{gathered}
Z_{eeg} = \text{LayerNorm}(X_{eeg} + \text{MHA}(Q_{E}, K_{S}, V_{S})) \\[6pt]
Z_{speech} = \text{LayerNorm}(X_{speech} + \text{MHA}(Q_{S}, K_{E}, V_{E})).
\end{gathered}
\end{equation}
The resulting outputs $Z_{eeg}$ and $Z_{speech}$ serve as inputs to both the GAM and a feed-forward network (FFN). Following the exact design of the FFN sublayer in vanilla transformers \cite{vaswani2017attention}, we apply two linear layers with ReLU activation independently to each modality:
\begin{equation}
\begin{gathered}
     X_{e} = \text{FFN}_e(Z_{eeg}), \quad X_{sp} = \text{FFN}_{sp}(Z_{speech}).
\end{gathered}
\end{equation}
The resulting outputs $X_{e}$ and $X_{sp}$, representing the refined EEG and speech embeddings, are combined with the GAM output to produce the final fused representations, as described in the Gating Alignment Module (Subs.~\ref{subsec:gam}).

\subsection{Gating Alignment Module}
\label{subsec:gam}
Inspired by the attention gating mechanism in \cite{qiu2025gated}, we propose a cross-modal Gating Alignment Module (GAM), shown in Fig.~\ref{fig:framework}(b). GAM selectively fuses speech and EEG features for MER. Given $X_{speech}$ and $X_{eeg}$, we first concatenate both modalities along the time dimension, and apply self-attention to the joint sequence to enable direct cross-modal interaction between speech and EEG embeddings. This is formulated below:
\begin{equation}
\begin{gathered}
    % Z_{\text{concat}} = \text{Concat}(Z_{eeg},\ Z_{speech}) \\[6pt]
    Z_{joint} = \text{SelfAtt}(\text{Concat}(Z_{eeg},\ Z_{speech})),
\end{gathered}
\end{equation} where $Z_{joint} \in \mathbb{R}^{2T \times F}$ is the combined representation of EEG and speech, capturing feature interactions across both modalities. We then pass through a simple gating mechanism to suppress emotionally irrelevant information. We pass $\hat{Z}$ through a sigmoid-gated bottleneck as shown below:
\begin{equation}
\begin{gathered}
    h = \sigma(W_{d}Z_{joint} + b_{d}) \\[6pt]
    Z_{gated} = W_{u} h + b_{u},
\end{gathered}
\end{equation}
where $W_{d}$ compresses the feature dimension to $F/2$, and $W_{u}$ projects back to the original dimension $F$. This yields the gated embedding $Z_{gated} \in \mathbb{R}^{2T \times F}$. The bottleneck compression reduces redundancy across both modalities as shown in \cite{nagrani2021attention}, while the sigmoid gate provides soft feature selection over the cross-modal interactions. $Z_{gated}$ is then split back into speech and EEG streams by dividing the temporal dimension in half, yielding $\hat{Z}_{speech} \in \mathbb{R}^{T \times F}$ and $\hat{Z}_{eeg} \in \mathbb{R}^{T \times F}$, matching the original dimensions of $Z_{speech}$ and $Z_{eeg}$ prior to GAM.  We finally add  $\hat{Z}_{speech}$ and $\hat{Z}_{eeg}$ to each modality to preserve modality-specific features while incorporating shared cross-modal context:
\begin{equation}
    X_{speech}' = X_{sp} + \hat{Z}_{speech}, \quad X_{eeg}' = X_{e} + \hat{Z}_{eeg}.
\end{equation} In general, this ensures that modality-specific features are preserved before passing into the \textit{Final Classification Layer} to output the emotion logit.

\subsection{Final Classification Layer}
The \textit{Final Classification Layer} serves as the decision module that maps $\hat{Z}_{speech}$ and $\hat{Z}_{eeg}$ to a predicted emotion. Both representations are first concatenated along the feature dimension to form a unified multimodal representation, which is then averaged across the time dimension through mean pooling to produce $X_{pool} \in \mathbb{R}^{F}$, a fixed-size summary of the joint representation. $X_{pool}$ is then passed through a linear layer with batch normalization and another linear layer to produce the final emotion prediction. This is displayed below:
\begin{equation}
\begin{gathered}
    X_{pool} = \text{MeanPool}(\text{Concat}(X_{speech}', \ X_{eeg}') \\[6pt]
    X_{hide} = \text{ReLU}(W_{hide} X_{pool} + b_h) \\[6pt]
    Y_{out} = \text{Softmax}(W_{out} X_{hide} + b_{out}),
\end{gathered}
\end{equation} where $W_h$ and $W_{out}$ represents the learnable weight matrices, and $\text{ReLU}$ is the activation function for the hidden layer. The final output $Y_{out} \in \mathbb{R}^{N}$ produces the predicted emotion class over $N$ emotion categories.

\begin{table*}[t]
\centering
\small
\caption{Comparison of EmoSpeechBrain against baseline and state-of-the-art EEG encoders on the PME4 and EAV datasets. We use $^*$ to indicate statistically significant improvement over all baselines (two-tailed paired $t$-test, $p < 0.05$).}
\begin{tabular}{c|cc|cc}
\hline
\multirow{2}{*}{\centering\textbf{Method}}
& \multicolumn{2}{c|}{\textbf{PME4}}
& \multicolumn{2}{c}{\textbf{EAV}} \\
& \textbf{Balanced Acc.} $\uparrow$
& \textbf{Weighted F1} $\uparrow$
& \textbf{Balanced Acc.} $\uparrow$
& \textbf{Weighted F1} $\uparrow$ \\
\cline{2-5}
\hline
\\[-1.8ex]
EEGNet \cite{lawhern2018eegnet}
& $0.503 \pm 0.035$
& $0.492 \pm 0.046$
& $0.551 \pm 0.111$
& $0.556 \pm 0.101$ \\[0.2ex]
BENDR \cite{kostas2021bendr}
& $0.543 \pm 0.065$
& $0.522 \pm 0.059$
& $0.646 \pm 0.015$
& $0.656 \pm 0.017$ \\[0.2ex]
BIOT \cite{yang2023biot}
& $0.522 \pm 0.071$
& $0.536 \pm 0.069$
& $0.635 \pm 0.023$
& $0.639 \pm 0.019$ \\[0.2ex]
EEGPT \cite{wang2024eegpt}
& $0.557 \pm 0.058$
& $0.548 \pm 0.061$
& $0.713 \pm 0.072$
& $0.718 \pm 0.077$ \\[0.2ex]
LaBraM \cite{jiang2024labram}
& $0.587 \pm 0.562$
& $0.594 \pm 0.052$
& $0.730 \pm 0.082$
& $0.733 \pm 0.081$ \\[0.2ex]
CBraMod \cite{wang2025cbramod}
& $0.605 \pm 0.072$
& $0.584 \pm 0.088$
& $0.736 \pm 0.078$
& $0.735 \pm 0.074$ \\[0.4ex]
\hline
\\[-1.8ex]
\textbf{EmoSpeechBrain (ours)}
& $\mathbf{0.734 \pm 0.015}^*$
& $\mathbf{0.709 \pm 0.014}^*$
& $\mathbf{0.799 \pm 0.043}^*$
& $\mathbf{0.785 \pm 0.067}^*$\\
\hline
\end{tabular}
\label{tab:results}
\end{table*}

\begin{table*}[t]
\centering
\small
\caption{Ablation study table measuring the accuracy of different model variants to evaluate the contribution of each component. We use $^*$ to indicate statistically significant improvement over all baselines (two-tailed paired $t$-test, $p < 0.05$).}
\begin{tabular}{c|cc|cc}
\hline
\multirow{2}{*}{\raisebox{-0.1ex}{\textbf{Method}}}
& \multicolumn{2}{c|}{\textbf{PME4}}
& \multicolumn{2}{c}{\textbf{EAV}} \\
& \textbf{Balanced Acc.} $\uparrow$
& \textbf{Weighted F1} $\uparrow$
& \textbf{Balanced Acc.} $\uparrow$
& \textbf{Weighted F1} $\uparrow$ \\
\cline{2-5}
\hline
\\[-1.8ex]
\textit{\textbf{Ablation on Each Modality}} \\[0.4ex]
\quad Speech Only
& $0.603 \pm 0.022$ & $0.594 \pm 0.024$
& $0.642 \pm 0.012$ & $0.621 \pm 0.013$ \\[0.2ex]
\quad EEG Only
& $0.503 \pm 0.030$ & $0.502 \pm 0.031$
& $0.562 \pm 0.032$ & $0.565 \pm 0.032$ \\[0.4ex]
\hline
\\[-1.8ex]
\textit{\textbf{Ablation on Different Components}} \\[0.4ex]
\quad w/o Local Feature Extraction
& $0.699 \pm 0.021$ & $0.693 \pm 0.022$
& $0.808 \pm 0.083$ & $0.807 \pm 0.063$ \\[0.2ex]
\quad w/o Differential Attention
& $0.653 \pm 0.023$ & $0.637 \pm 0.025$
& $0.756 \pm 0.052$ & $0.756 \pm 0.049$ \\[0.2ex]
\quad w/o GAM
& $0.621 \pm 0.023$ & $0.619 \pm 0.024$
& $0.742 \pm 0.049$ & $0.727 \pm 0.055$ \\[0.2ex]
\quad w/o Diff. Attn + w/o GAM
& $0.613 \pm 0.012$ & $0.609 \pm 0.022$
& $0.721 \pm 0.044$ & $0.718 \pm 0.069$ \\[0.4ex]
\hline
\\[-1.8ex]
\textbf{EmoSpeechBrain (ours)}
& $\mathbf{0.734 \pm 0.015}^*$ & $\mathbf{0.709 \pm 0.014}^*$
& $\mathbf{0.799 \pm 0.043}^*$ & $\mathbf{0.785 \pm 0.067}^*$
\\
\hline
\end{tabular}
\label{tab:ablation1}
\end{table*}

\section{Experimental Setup}
\subsection{Dataset Selection}
The experiments for EmoSpeechBrain were conducted on two datasets: PME4 \cite{pme4dataset} and EAV \cite{lee2024eav}, which provide paired EEG and speech recordings for multimodal emotion recognition. We summarize the details of the datasets below.
%\shr{I find that the paragraph you have commented out here is critical - it is always important to justify WHY you chose these datasets}

\textbf{PME4 Dataset \cite{pme4dataset}:} PME4 is a multimodal emotion dataset containing audio, video, EEG, and EMG recordings from 11 acting students (5 females and 6 males). EEG was recorded at 5 kHz and downsampled to 1 kHz, with audio at 44.1 kHz, covering 7 emotion categories: anger, fear, disgust, sadness, happiness, surprise, and neutral.

\textbf{EAV Dataset \cite{lee2024eav}:} EAV is a multimodal emotion dataset designed for emotion recognition in conversational contexts. It contains 30-channel EEG recordings sampled at 500 Hz alongside synchronized audio and video from 42 subjects. Each subject completes 200 interactions in a cue-based dialogue, where every interaction pairs a 20-second listening segment with a 20-second speaking segment, covering 5 emotion categories: neutral, anger, happiness, sadness, and calmness. Since audio recordings are only available for the speaking segments (no audio is captured during listening), we restrict our experiments to the speaking-only portion of the data, yielding 100 trials per subject with EEG, audio, and video jointly available.

\subsection{Dataset Preprocessing}
\textbf{EEG Preprocessing:} Raw EEG signals were bandpass filtered between 1.0 and 40.0 Hz using a zero-phase 4th-order Butterworth filter, downsampled to 200 Hz, and segmented into non-overlapping 2-second epochs. Each epoch was independently normalized per channel using z-score standardization. All available channels were retained. \vspace{6pt}

\noindent \textbf{Speech Preprocessing:} Raw audio waveforms were resampled to 16 kHz (mono) and segmented into 2-second epochs to match the EEG segments. Each epoch was normalized by its peak absolute value to ensure a consistent amplitude scale.

\subsection{Training Configurations}
\begin{figure*}[t]
    \centering
    \includegraphics[width=\textwidth]{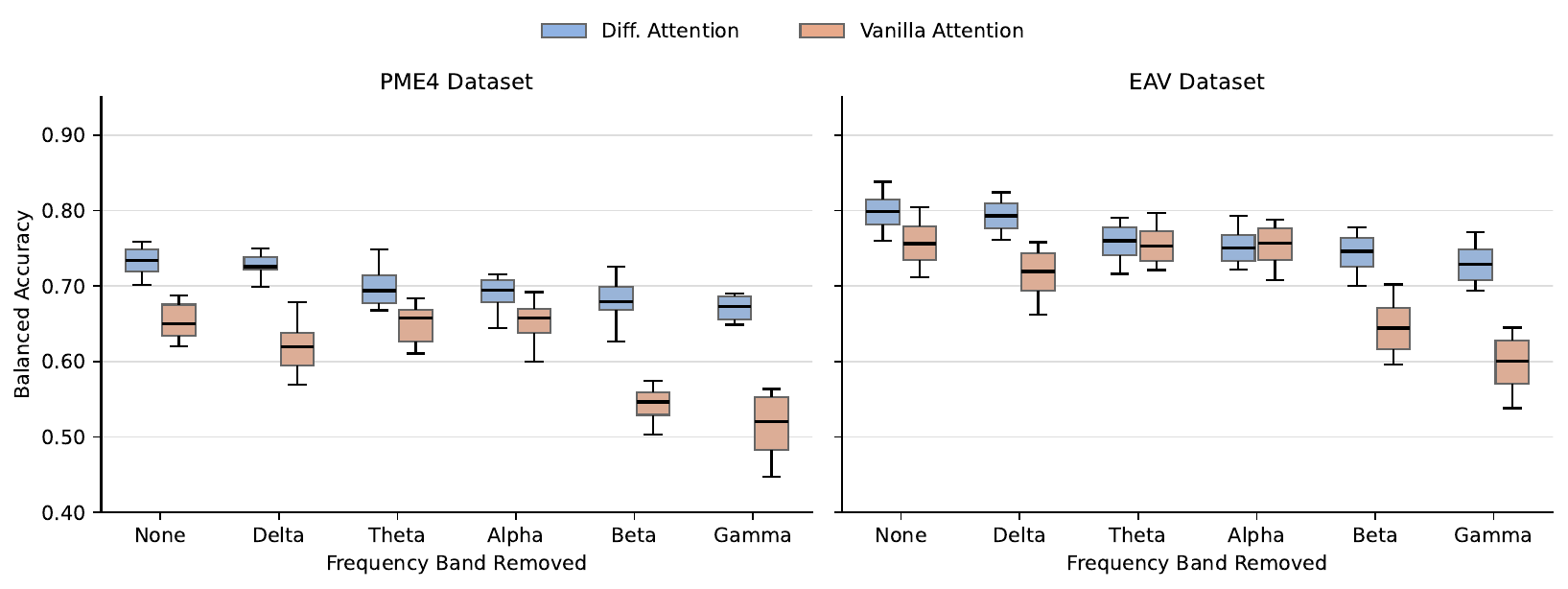}
    \caption{Ablation study showing balanced accuracy of differential attention and vanilla attention on PME4 and EAV when each EEG frequency band is removed. \textit{None} denotes the EEG has no frequency bands removed.}
    \label{fig:freq_ablation}
\end{figure*}

We evaluate all models using a leave-one-subject-out (LOSO) protocol, where we hold out each subject for testing and train the model on the remaining subjects. For each test subject, we compute accuracy and F1-score and then average them across all subjects to obtain the final results. We use this same protocol and these same test subjects for the main results in Table~\ref{tab:results}, the component ablation in Table~\ref{tab:ablation1}, and the frequency ablation in Fig.~\ref{fig:freq_ablation}.

In the main results, we compare EmoSpeechBrain against each baseline method. In the component ablation, we compare EmoSpeechBrain against each variant with an individual component removed. To assess significance, we pair accuracies on every test subject and apply a two-tailed paired $t$-test, following~\cite{dietterich1998approximate}. For the main results, we test the significance testing of EmoSpeechBrain against all of the baselines. For the component ablation, we test the significance testing of EmoSpeechBrain against each variant. An $^{\ast}$ on EmoSpeechBrain's result indicates a statistically significant difference ($p < 0.05$) from all compared models.

For the frequency ablation, we create five variations of each test subject's data, each with a single band removed by a band-stop filter: \textit{delta} (1--4~Hz), \textit{theta} (4--8~Hz), \textit{alpha} (8--13~Hz), \textit{beta} (13--30~Hz), or \textit{gamma} (30--45~Hz). We evaluate the trained model on each variation and on the original test data with \textit{none} of the bands removed, without any retraining. We then report the relative drop in balanced accuracy of each variation with respect to the \textit{none} result, averaged across all test subjects.

EmoSpeechBrain is trained using the Adam optimizer with a constant learning rate of $1 \times 10^{-5}$, a weight decay of $3\times10^{-4}$, and a batch size of 8 for up to 50 epochs with early stopping (patience of 5 epochs) on a single NVIDIA RTX 4090 GPU, while all baselines are trained following their original setups.

\subsection{Baselines}
We evaluate EmoSpeechBrain against task-specific architectures and large-scale pretrained EEG foundation models to assess whether standard EEG encoders or large-scale EEG pretraining alone is sufficient for MER.

\section{Results}
\subsection{Model Performance}
Table~\ref{tab:results} presents a comparison of EmoSpeechBrain against competitive EEG encoder baselines spanning both task-specific architectures and large-scale EEG foundation models. For a fair comparison, all baselines utilize the same frozen WavLM-Large speech encoder, with only the EEG encoder and cross-modal fusion components left unfrozen during training.

EmoSpeechBrain achieves the highest performance across both datasets. In PME4, it achieves a balanced accuracy of $0.734 \pm 0.012$ and a weighted F1 of $0.709 \pm 0.014$, surpassing the strongest baseline CBraMod \cite{wang2025cbramod} by over 10\%. On EAV, it achieves $0.799 \pm 0.043$ balanced accuracy and $0.785 \pm 0.067$ weighted F1, exceeding the best competing method by a clear margin. All improvements are statistically significant over every baseline ($^{\ast}p < 0.05$, two-tailed paired $t$-test).

In general, the results in Table~\ref{tab:results} demonstrate that both task-specific EEG encoders and foundation models consistently underperform compared to EmoSpeechBrain in both datasets. This suggests that neither standard EEG encoding nor large-scale EEG pretraining alone is sufficient for MER, as these models lack an explicit mechanism to bridge EEG and speech representations. EmoSpeechBrain addresses this through its EEG Encoder and GAM, which jointly shape EEG representations toward the speech feature space during fine-tuning, yielding more discriminative cross-modal representations.

\section{Ablation Studies}
\subsection{Modality Ablation}
To validate the necessity of both modalities, we evaluate two unimodal variants with the majority of the configurations set identical to our proposed method. For the speech-only configuration, we retain the frozen WavLM-Large encoder with all remaining components unfrozen, replacing the cross-modal fusion with a single-modality classification head. For the EEG-only configuration, we retain the full proposed EEG encoder architecture but remove the speech branch and fusion layer, feeding EEG embeddings directly to the classifier.

As shown in the modality ablation section of Table~\ref{tab:ablation1}, neither modality alone is sufficient for effective emotion recognition. The speech-only modality achieves $0.603 \pm 0.022$ accuracy on PME4 and $0.642 \pm 0.010$ accuracy on EAV, while the EEG-only modality drops further to $0.503 \pm 0.030$ accuracy on PME4 and $0.562 \pm 0.032$ accuracy on EAV. Both unimodal variants show statistically significant degradation compared to the full model, confirming that neural and acoustic signals carry complementary emotional information, and that combining both modalities is essential to find complementary information in emotion recognition.

\subsection{Architectural Ablation}
To evaluate the individual contribution of each proposed architectural component in EmoSpeechBrain, we examine four configurations: (1) removing local feature extraction layer, (2) replacing differential attention with standard multi-head attention, (3) removing the GAM, and finally, (4) removing both the GAM and differential attention simultaneously. For each configuration, all other settings are identical to the full model.

As shown in the component ablation section of Tables~\ref{tab:ablation1}, removing any component consistently degrades both classification performance and cross-modal alignment quality, with all differences statistically significant compared to the full model. Removing local feature extraction causes a marginal drop ($0.699 \pm 0.021$ on PME4), while replacing differential attention with standard multi-head attention produces a more notable decline ($0.653 \pm 0.023$ on PME4, $0.756 \pm 0.052$ on EAV), validating that noise suppression is critical to obtain relevant information from EEG signals. The largest single-component drop comes from removing the GAM, reducing accuracy to $0.621 \pm 0.023$ on PME4 while EAV, accuracy reduces to to $0.742 \pm 0.049$. This confirms that explicit cross-modal fusion is essential for grounding EEG representations with speech features. Removing both the GAM and differential attention simultaneously yields the worst overall results ($0.613 \pm 0.012$ accuracy on PME4, $0.721 \pm 0.044$ accuracy on EAV), underscoring the compounding benefit of both components. These results collectively demonstrate that effective MER requires both a well-designed EEG encoder that captures meaningful emotional representations rather than spurious artifacts, and an explicit fusion mechanism to ensure those representations are properly grounded with speech features.

\subsection{Frequency Ablation}
Prior work shows that \textit{beta} and \textit{gamma} are the most useful bands for EEG-based emotion recognition \cite{zheng2015investigating, jenke2014feature}. However, involuntary movement heavily contaminates EEG above 20~Hz \cite{whitham2007scalp}, so a model that relies on these bands may partly be learning from noise. Lower bands also carry emotional information. For example, frontal \textit{alpha} asymmetry is linked to emotional valence \cite{coan2004frontal}. Based on these findings, we overall hypothesize that vanilla attention relies mostly on the noisy high-frequency bands, while differential attention cancels shared noise and uses the full spectrum more evenly.

As shown in Fig. \ref{fig:freq_ablation}, removing \textit{beta} and \textit{gamma} causes the largest drop for both models. However, vanilla attention drops much more (15.5--21.1\% on PME4 and 14.8--20.6\% on EAV) than differential attention (7.2--9.1\% and 6.6--8.8\%). In contrast, removing \textit{alpha} and \textit{theta} barely affects vanilla attention (at most 0.5\% on both datasets) but clearly lowers differential attention (4.8--6.0\% on PME4 and 4.9--6.1\% on EAV). This suggests that vanilla attention depends mostly on noisy high-frequency bands, while differential attention also uses lower-frequency bands. Since speech can captures arousal better than valence \cite{wagner2023dawn}, this wider use of EEG may add information that speech lacks, which is consistent with gains of differential attention.

\section{Limitations}
While EmoSpeechBrain outperforms SOTA EEG foundation models and unimodal baselines, our comparisons focus mainly on the EEG encoder. All baselines in Table~\ref{tab:results} are EEG encoders evaluated within our framework, and we use the same speech encoder across all of them. We keep the speech encoder fixed so that differences in performance come from the EEG encoder alone. These results show that our EEG encoder produces representations better suited for fusion with speech, but not that our fusion strategy outperforms prior EEG--speech fusion methods. We have not yet directly compared EmoSpeechBrain against existing EEG--speech fusion systems, as prior work in this area is limited and often differs in datasets, preprocessing, and evaluation protocols. In future work, we plan to evaluate different speech encoders and fusion methods under the same LOSO protocol to separately assess the contribution of each component.

\section{Conclusion}
In this paper, we propose EmoSpeechBrain, a multimodal framework that addresses the challenge of effectively combining EEG signals with speech for accurate MER. Through our proposed differential attention-based EEG encoder and GAM, EmoSpeechBrain achieves SOTA MER accuracy across two datasets while significantly outperforming unimodal baselines. Ablation results further confirm that both components are necessary, demonstrating that effective MER requires information-rich EEG representations and an explicit mechanism to bridge both modalities together.

\section{Generative AI Use Disclosure}
Generative AI tools were used exclusively to refine the language of author-written text. No scientific content, results, experimental designs, analyses, or conclusions were generated using these tools. All authors take full responsibility for the paper's content and consent to its submission.

\bibliography{refs}
\bibliographystyle{IEEEtran}
\end{document}